\def\year{2020}\relax
\documentclass[letterpaper]{article} 
\usepackage{aaai20}  
\usepackage{times}  
\usepackage{helvet} 
\usepackage{courier}  
\usepackage[hyphens]{url}  
\usepackage{graphicx} 
\usepackage{graphicx}  
\usepackage{epsfig}
\usepackage{amsmath}
\usepackage{amssymb}
\usepackage{float}
\usepackage{subcaption} 

\makeatletter
    \renewcommand{\copyright@on}{F}
    \AtEndDocument{%
    }
\makeatother

\newcommand{\name}{RoadTagger}

\title{\name: Robust Road Attribute Inference with Graph Neural Networks}
\author{
Songtao He,\textsuperscript{\rm 1}
Favyen Bastani,\textsuperscript{\rm 1}
Satvat Jagwani,\textsuperscript{\rm 1}
Edward Park,\textsuperscript{\rm 1} 
Sofiane Abbar,\textsuperscript{\rm 2} \\
\bf \Large Mohammad Alizadeh,\textsuperscript{\rm 1}
Hari Balakrishnan,\textsuperscript{\rm 1}
Sanjay Chawla,\textsuperscript{\rm 2} \\
\bf \Large  Samuel Madden,\textsuperscript{\rm 1}
Mohammad Amin Sadeghi\textsuperscript{\rm 3}\\
\textsuperscript{\rm 1} MIT Computer Science and Artificial Intelligence Laboratory (MIT CSAIL)\\
\textsuperscript{\rm 2} Qatar Computing Research Institute (QCRI)\\
\textsuperscript{\rm 3} University of Tehran \\
songtao@csail.mit.edu, favyen@csail.mit.edu, satvat@csail.mit.edu, parke@csail.mit.edu, sabbar@hbku.edu.qa,\\ alizadeh@csail.mit.edu, hari@csail.mit.edu, schawla@hbku.edu.qa,\\ madden@csail.mit.edu, msadegh2@illinois.edu
}

\begin{document}
\maketitle
\begin{abstract}

Inferring road attributes such as lane count and road type from satellite imagery is challenging. Often, due to the occlusion in satellite imagery and the spatial correlation of road attributes, a road attribute at one position on a road may only be apparent when considering far-away segments of the road. Thus, to robustly infer road attributes, the model must integrate scattered information and capture the spatial correlation of features along roads. Existing solutions that rely on image classifiers fail to capture this correlation, resulting in poor accuracy. We find this failure is caused by a fundamental limitation -- the limited effective receptive field of image classifiers. 

To overcome this limitation, we propose \name\footnote{The source code of \name\ project is available at \texttt{https://github.com/mitroadmaps/roadtagger.git}},  an \textit{end-to-end} architecture which combines both Convolutional Neural Networks (CNNs) and Graph Neural Networks (GNNs) to infer road attributes. Using a GNN allows information to propagate on the road network graph and eliminates the receptive field limitation of image classifiers. We evaluate \name\ on both a large real-world dataset covering 688 $km^2$ area in 20 U.S. cities and a synthesized dataset. In the evaluation, \name\ improves inference accuracy over the CNN image classifier based approaches. In addition, \name\ is robust to disruptions in the satellite imagery and is able to learn complicated inductive rules for aggregating scattered information along the road network.

\end{abstract}

\section{Introduction}
Detailed road attributes enrich maps and enable numerous new applications. For example, mapping the number of lanes on each road makes lane-to-lane navigation possible, where a navigation system informs the driver which lanes will diverge to the correct branch at a junction. Similarly, maps that incorporate the presence of bicycle lanes along each road enable cyclists to make more informed decisions when choosing a route. Additionally, maps with up-to-date road conditions and road types improve the efficiency of road maintenance and disaster relief. 

Unfortunately, producing and maintaining digital maps with road attributes is tedious and labor-intensive. In this paper, we show how to automate the inference of road attributes from satellite imagery.

Consider the problem of determining the number of lanes on the road from images. A natural approach would be to map this problem to an image classification problem. Because the number of lanes of one road may vary, we can scan the road with a sliding window and train a classifier to predict the number of lanes in each window along the road independently. After we have the classifier predictions in each window, we can apply a post-processing step to improve the prediction results; e.g., using the road network graph to remove inconsistent predictions along the road. Some prior map inference papers adopt such a strategy~\cite{cadamuro2018assigning,najjar2017combining}.

\begin{figure}[t]
	\begin{center}
		\includegraphics[width=\linewidth]{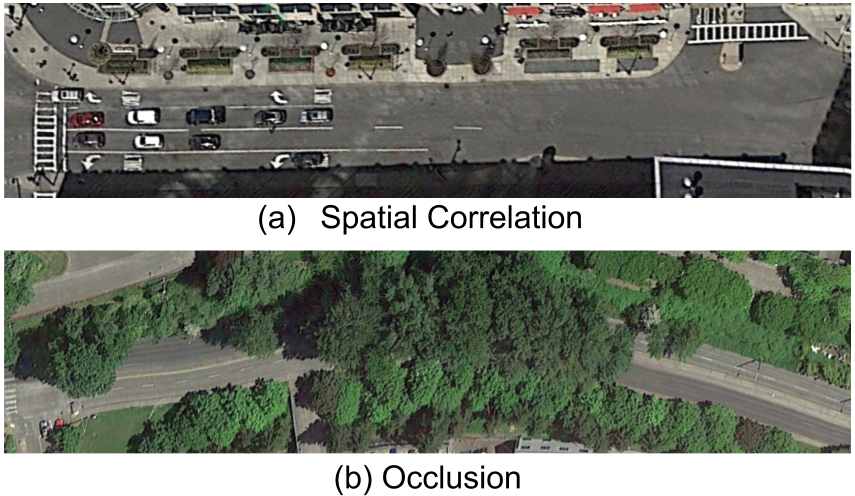}
	\end{center}

	\caption{Challenges in road attribute inference. In (a) the lane markings are absent on one side of the road. In (b) the road is occluded by trees and the lane markings are also partially missing. To make correct predictions at all positions on the road, we need to incorporate both the local information and the global information along the road network graph. }
	
	\label{fig:introExample}
\end{figure}

\begin{figure*}[t]
	\begin{center}
		\includegraphics[width=\linewidth]{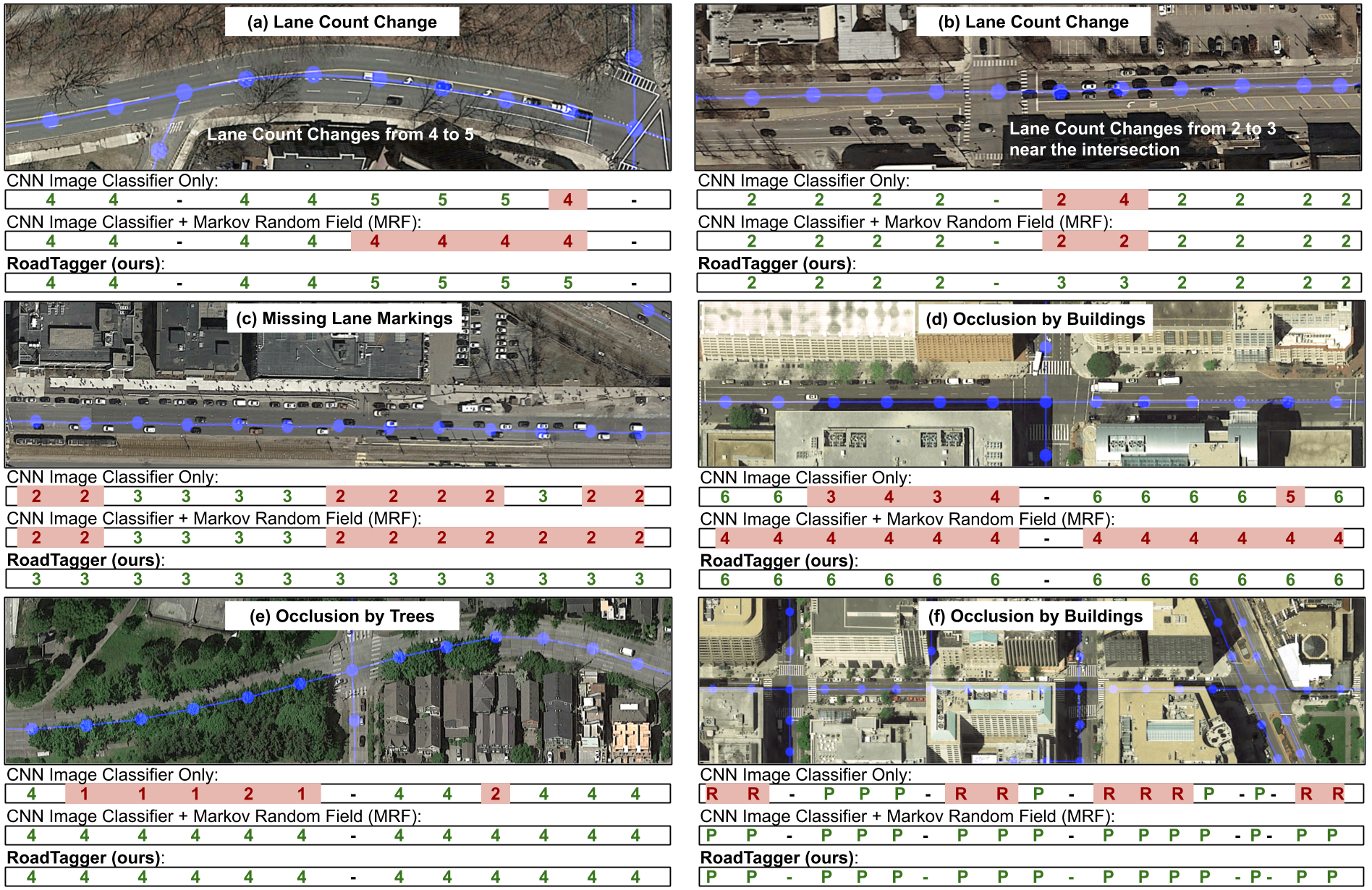}
	\end{center}
	\caption{Examples of lane inference (examples (a-e)) and road type inference (example (f)). In each image, blue lines show the road graph. The number of lanes or the type of the road predicted by the CNN Image Classifier (with and without MRF) and \name\ on each segment are shown along the bottom of each figure. For road type inference, we use capital P to represent primary roads and capital R to represent residential roads. We color the output numbers and letters green for correct predictions and red for incorrect predictions.}

	\label{fig:DeepRoadGraphExample}
\end{figure*}

This approach suffers from a fundamental limitation: the \textit{limited effective-receptive-field of image classifiers.} Consider Figure \ref{fig:introExample}(a), where lane markings are visible only on the left. Because the road width remains the same, these lane markings imply the remainder of the road on the right has the same number of lanes, despite not having explicit markings. However, because practical image classifiers can only be scalably trained with small windows of the city-wide satellite imagery as input, a window-based image classifier cannot capture this spatial correlation and would not correctly predict the number of lanes in the right portion of the road. Thus, the limited effective-receptive-field of the classifier does not capture the long-term spatial propagation of the image features needed to accurately infer road attributes. 

To overcome this limitation, prior work~\cite{mattyus2015enhancing} proposes adding a global inference phase to post-process the output from the local classifiers. We find that this fix is inadequate. For example, see Figure \ref{fig:DeepRoadGraphExample}(a), where the lane count changes from 4 to 5 near an intersection. The image classifier outputs partially incorrect labels. However, the post-processing strategy (Markov Random Field) cannot fix this problem as the global inference phase only takes the predictions from the image classifier as input and it may not be able to tell whether the number of lanes indeed changes or it is an error of the image classifier. This limitation is caused by the \textit{information barrier} induced by the separation of local classification and global inference; the global inference phase can only use the image classifier's prediction as input, but not other important information such as whether trees occlude the road or whether the road width changes.

We propose \name, an end-to-end road attribute inference framework that eliminates this barrier using a novel combination of a Convolutional Neural Network (CNN) and a Graph Neural Network (GNN)~\cite{wu2019comprehensive}. It takes both the satellite imagery and the road network graph
 as input. For each vertex in the road network graph, \name\ uses a CNN to derive a feature vector from a window of satellite imagery around the vertex. Then, the information from each vertex is propagated along the road network graph 
using a GNN. Finally, it produces the road attribute prediction at each vertex. The GNN eliminates the effective-receptive-field limitation of local image classifiers by propagating information along the road network graph. 
The end-to-end training of the combined CNN and GNN model is the key to the success of the method: \name\ doesn't select features using only the CNN; instead, by backpropagating from the output of the GNN, the information barrier that limited previous post-processing methods is eliminated. 

We  evaluate the performance and robustness of \name\  with both a real-world dataset covering a $688~\mathtt{km}^2$ area in 20 U.S. cities and a synthesized dataset focused on different challenges in road attribute inference. We focus on two types of road attributes: the number of lanes and the type of road (e.g., primary or residential). 
In the real-world dataset, we show that \name\ surpasses a set of CNN-based image classifier baselines (with and without post-processing). Compared with the CNN image classifier baseline, \name\ improves the inference accuracy of the number of lanes from 71.8\% to 77.2\%, and of the road type from 89.1\% to 93.1\%. This improvement comes with a reduction of the absolute lane detection error of 22.2\%. We show output examples in Figure \ref{fig:DeepRoadGraphExample}. On the synthesized dataset, we found that \name\ is able to learn complicated inductive rules and is robust to different disruptions.

\section{Related Work}
Cadamuro et al. adapt CNN image classifiers to predict road quality from imagery~\cite{cadamuro2018assigning}. Najjar et al. use satellite imagery to create a road safety map of a city~\cite{najjar2017combining} by adapting a CNN image classifier to assign a safety score to each input satellite image window. Azimi et al. apply a CNN to perform a semantic segmentation for lane markings~\cite{azimi2018aerial}. However, because these schemes derive labels directly from the CNN, they are only able to infer attributes that pertain to small objects (e.g. a satellite image window or a lane marker), and not attributes over an entire road.

To address this issue, M\'attyus et al. propose modeling the problem as an inference problem in a Markov Random Field (MRF)~\cite{mattyus2015enhancing}. They develop an MRF model that encodes low-level image features such as edge, pixel intensity, and image homogeneity; high-level image features such as road detector results and car detector results; and domain knowledge such as the smoothness of the road and overlapping constraints. They show that this model can infer various road attributes, including road width, centerline position, and parking lane location. They further extend the approach to take ground images (dashcam) into account~\cite{mattyus2016hd}. In contrast to this MRF approach, \name\ does not require specification of different features and domain knowledge for each road attribute. Instead, \name\ extracts the useful image features and domain knowledge through end-to-end learning, making it a powerful and easy-to-use framework for different road attributes.

Recent work has also explored using satellite imagery for fully automated road network inference and building footprint mapping. DeepRoadMapper~\cite{mattyus2017deeproadmapper} segments the satellite imagery to classify each pixel as road or not road. It then extracts the road network from the segmentation result and applies additional post-processing heuristics to enhance the road network. RoadTracer~\cite{bastani2018roadtracer} trains a CNN model to predict the direction of the road. It starts from a known location on the road and traces the road network based on the direction prediction from the CNN model. Hamaguchi et al.~\cite{hamaguchi2018building} use an ensemble of size-specific CNN building detectors to produce accurate building footprints for a wide range of building sizes. 

\begin{figure*}[t]
	\begin{center}
		\includegraphics[width=\linewidth]{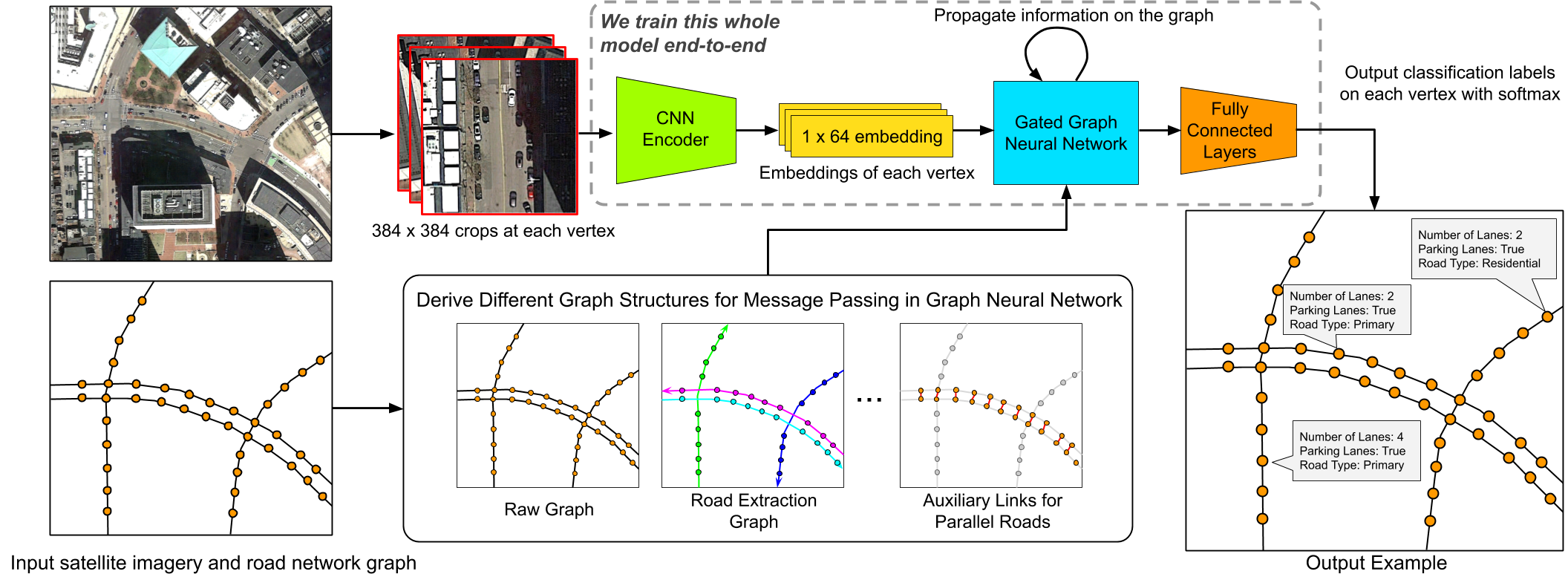}
	\end{center}
	\caption{The overview of \name\ road attribute inference framework.} 
	\label{fig:DeepRoadGraphArch}
\end{figure*}

\section{\name}
\name\ uses both a CNN and a Gated Graph Neural Network (GGNN)~\cite{li2015gated} to infer road attributes from satellite imagery and the corresponding road network graph. We assume here the road network graph is already available and accurate, since prior work shows how to infer the graph from satellite imagery~\cite{bastani2018roadtracer,mattyus2017deeproadmapper} or \label{method} GPS traces~\cite{gpsbiagioni,gpsahmed2015comparison,gpsedelkamp2003route,gpsdavies2006scalable,gpscao2009gps,gpskharita,gpsroadrunner}.

Figure \ref{fig:DeepRoadGraphArch} shows an overview of \name. The first step is to densify the road network graph so that there is one vertex every 20 meters. Then for each vertex in the graph, get the satellite imagery in the area around the vertex to a CNN encoder, and rotate it so that the road direction is always vertical in the image. Each image is $384 \times 384$ pixels, corresponding to a $48 \times 48$ meter tile at 12.5 cm/pixel image resolution. This high resolution is needed to capture details on the road such as lane markings. The CNN encoder uses 12 convolutional layers and 3 fully-connected layers to extract a 64-dimension embedding for each vertex. The method then passes the embeddings of all the vertices to the GNN module.  The GNN propagates local information around each vertex to neighbouring vertices on the road network graph. After a few steps of information exchanges, the GNN produces the final prediction of each vertex through three additional fully-connected layers and a soft-max layer. 

The model can be expressed as
\begin{equation}
    y_v = f_{\mbox{GNN}}(f_{\mbox{CNN}}(s_v), G)
\end{equation}
where $s_v$ is the input satellite image tile at vertex $v$, $f_{CNN}()$ is the CNN encoder, $G$ is the densified road network graph, $f_{GNN}()$ is the graph neural network module and $y_v$ is the set of output road attribute labels (soft-max) at vertex $v$.  We train the model end-to-end with cross-entropy loss using known ground-truth labels at each vertex $v$.

\subsection{Graph Neural Network Module}
A Gated GNN is an extension of a Recurrent Neural Network (RNN) on a graph, using a Gated Recurrent Unit (GRU)~\cite{chung2015gated} to  propagate information between adjacent vertices of the graph.  
We represent the embedding of the vertex $v$ as $x_v$. Before applying the GGNN, we extend the dimension of $x_v$ from 64 to 128 through two fully-connected layers (the $f_{\mbox{raise}}()$ function). Extending the dimension of the original embedding helps ensure that we don't induce an information bottleneck when the information is propagating on the graph. We represent the hidden state at propagation step $t$ of vertex $v$ as $h_v^t$. Then, the basic propagation model on the road network graph $\{V,E\}$ can be written as
\begin{equation}
\begin{aligned}
&h_v^0 = [f_{raise}(x_v)] \\
&m_v^t = f_1(h_v^{t-1})\\
&a_v^t = \frac{1}{|N(v)|}\sum_{u \in N(v)} m_u^t  \\
&h_v^t = f_{GRU}(h_v^{t-1}, a_v^t)\\
\end{aligned}
\label{equation:GRU}
\end{equation}
Here, 
$N(v)$ is a function representing the set of all logical neighbors of $v$ (explained below),
$f_1$ is a fully-connected layer, and $f_{GRU}$ is a GRU. The algorithm uses this propagation function to propagate the information on the graph for $T$ steps, finally producing the prediction labels through three additional fully-connected layers and a soft-max layer. 

\subsubsection{Graph Structures}
\label{graph_structure}
In the propagation model (\ref{equation:GRU}), the choice of edge placement defined by $N(v)$ is critical to the inference performance;
because it controls how the vertices in the graph communicate with each other. 

We investigate different {\em graph structures} derived from the original road network graph to define $N(v)$. These graph structures share the same set of vertices as the original road network graph but may have different edges defined by $N(v)$. Because the graph structure controls the propagation of information, we can use it to restrict communication to only a subset of the graph or enable communication between two vertices that were not connected in the original road network graph. A good graph structure can improve the performance of the graph neural network. We now discuss four example graph structures:

\begin{enumerate}
\item {\em Original Graph.} The structure enables communication between all connected vertices in the road network graph. It allows the GNN to learn the best way of communication without any restriction or preference. However, this freedom also makes it less efficient to learn certain types of road attributes; such as for road-specific attributes, when two roads with different road-specific attributes interact at an intersection, the graph neural network model need to make sure the information from one road won't mess up with the information from the other road. 

\item {\em Road Extraction Graph.} This structure helps propagate messages only within the same road. To automate this process, extract road chains belonging to the same logical road from the original graph. Here, we call a sequence of connected edges a ``road chain belonging to the same logical road'' when the directional difference between any two consecutive edges is less than 60 degrees. We show examples of road chains with different colors in Figure~\ref{fig:DeepRoadGraphArch}. We find that this restriction is helpful for the propagation of road-specific information as it removes the ambiguity at intersections.

\item {\em Road Extraction Graph (Directional).} We can decompose one road chain into two chains with opposing directions. This decomposition yields two separate graph structures. By providing two graph structures with opposite edge direction to the GNN, we can explicitly specify the source of each message. This modification can help the GNN learn more efficiently. 

\item {\em Auxiliary Graph for Parallel Road Pairs.} The original graph adapts two separate vertex chains to represent a parallel road pair. This  representation prevents communication between the two roads in a parallel road pair. We enable this communication by adding auxiliary edges between the two roads in a parallel road pair. We show an example of the auxiliary edges in Figure~\ref{fig:DeepRoadGraphArch}.
The intuition here is that roads belonging to the same parallel road pair often share the same road attributes, e.g., road type. If one road in a pair of parallel roads is occluded by buildings or trees in the satellite image, \name\ can still infer the road attributes correctly by incorporating information from the other road through the auxiliary edges.   

\end{enumerate}

We find all these graph structures improve \name\ in different ways. Thus, we extend the propagation model (\ref{equation:GRU}) to support multiple graph structures. Instead of aggregating messages from different graph structures together, we treat them separately. To support $k$ different graph structures, we extend the dimension of the hidden state from $m$ to $k\times m$, where the messages from the $i$-th graph structure are stored in the $i$-th $m$-dimensional chunk in the hidden state vector. Although messages from different graph structures are stored separately, they can still interact with each other in the GRU ($f_{GRU}()$).

\subsection{Training \name}
Training a deep model with both a CNN and a GNN is not straightforward. We found that training \name\ with standard cross-entropy loss and common anti-overfitting techniques such as data augmentation and dropout was insufficient. This is because, compared with CNN-based image classifiers on the same training dataset, \name\ usually has more parameters, but perceives less diversity.

Moreover, nodes in \name\ have a larger set of inputs than the local information in CNNs due to the usage of graph neural networks. 
This extra information makes \name\ even easier to overfit.  Unfortunately, there is no well-known regularization mechanism that can be applied during training procedure to prevent \name\  from overfitting to this extra information. 

In order to prevent \name\ from overfitting, we explore two training techniques, \textit{random vertex dropout} and \textit{graph Laplace regularization}~\cite{smola2003kernels}.

\textbf{Random Vertex Dropout.}
We represent the embedding of vertex $v$ as $e_v$. During training, we randomly pick up 10\% of vertices and set their embedding to a random vector $e_v^*$ where,
\begin{equation}
e_v^* = e_v \odot r,  \quad r_i \sim \mathcal{U}([-1,1])
\end{equation}
Here, we use $\odot$ to denote element-wise multiplication. We stop the gradient back-propagation for the dropped vertices.  This random vertex dropout is an extension of the standard dropout. However, instead of setting the embeddings to all zeros, we set them to a random vector. This is because an all-zeros vector is too easy for the neural network to distinguish.
We aim to use this random vertex dropout to simulate scenarios where the road is partially occluded by trees, buildings or bridges. This can increase the diversity of the training dataset and thus reduce over-fitting. 

\textbf{Graph Laplace Regularization.}
As we mentioned, there is no regulation mechanism in our training procedure to prevent \name\ from abusing the extra information. To overcome this limitation, we add a regularization term based on graph Laplace regularization to the final loss function. We represent the final soft-max output for vertex $v$ as a $n$-dimensional vector $\mathbf{y_v}$, where $n$ is the number of classes of the road attribute. Then, the regularization term for vertex $v$ can be written as, 
\begin{equation}
\mathtt{L}_{\mathtt{reg}}(v) = \lambda(v) |\mathbf{y_v} -   \frac{1}{|N(v)|} \sum_{u \in N(v)} \mathbf{y_u}|^2
\end{equation} 
where $\lambda$ is the weight of the regularization term at different vertices. We set $\lambda(v)$ to zero if vertex $v$ and its neighbours $N(v)$ have inconsistent ground truth labels. Otherwise, we set $\lambda(v)$ to a constant weight factor. 

This additional term forces \name\ to generate consistent labels for neighbouring vertices regardless of their correctness. It acts as a regularization term for the cross-entropy loss, which only focuses on per-vertex correctness; thus, it reduces overfitting.

\section{Evaluation}
In this section, we evaluate the performance and robustness of \name. We compare \name\ against CNN image classifier based solutions. In the evaluation, we focus on the architecture comparison between \name's end-to-end CNN+GNN framework and the CNN only image classifier solution. We use the same configuration for all the convolutional layers and fully connected layers except the last one in both \name's CNN encoder and the CNN image classifier. We also use the same input satellite image size for both \name\ and the CNN image classifier. 

In the evaluation, we demonstrate \name's performance improvement in road attribute inference in a large scale real world environment. We evaluated the performance of different variants of \name\ . In addition, we show when and why \name\ can yield better performance and analyze its limitations via a robustness study on a synthesized dataset. 

\subsection{Dataset}
We conduct our evaluation on two datasets, one real-world dataset and one synthetic micro-benchmark. For the real-world dataset, we collect the road attributes (ground truth labels) from OpenStreetMap~\cite{haklay2008openstreetmap} and the corresponding satellite imagery through the Google static map API~\cite{googleapi}. This dataset covers 688 $km^2$ area in 20 U.S. cities. We manually verified the labels of 16 $km^2$ of the dataset from four representative cities: Boston, Chicago, Washington D.C., and Seattle. We use one third of the verified dataset as validation dataset and two thirds of it as testing dataset. We use all the remaining dataset as training dataset.

We focus on inferring two types of road attribute: the number of lanes and the types of roads (residential roads or primary roads). We use these two types of road attributes as representatives because they both have spatial correlation such that nearby segments tend to have the same labels.

\subsection{Implementation Details}
We implemented both \name\ and the CNN image classifier using Tensorflow~\cite{abadi2016tensorflow}. We use both input image augmentation and dropout to reduce over-fitting. For \name, we set the graph Laplace regularization weight to $3.0$ and set the propagation step to 8 in the graph neural network. We train the model to predict both the number of lanes and the type of the road simultaneously. We train the model on a V100 GPU for 300k iterations with a learning rate starting from 0.0001 and decreasing by 3x every 30k iterations. For both models, we use a batch size of 128. In \name, at each iteration, we pick up a random vertex in the road network graph and start a DFS or BFS from it. We use the first 256 vertices in the search result to generate a sub-graph as input graph to \name. We only consider a random 128 vertices from the 256 vertices in the loss function.  For both models, we use batch normalization to speed up training. We use the model that performs best on the validation set as the final model.

\subsection{Baselines}
We compare \name\ against four different baselines, including (1) using only CNN image classifier, (2) using CNN image classifier with smoothing post-processing, (3) using CNN image classifier with Markov Random Field (MRF) post-processing, and (4) using CNN image classifier with larger receptive fields (1.5x and 2.0x).

In the smoothing post-processing approach, we set the probability outputs of each vertex to be the average probability of itself and its neighbouring vertices in the road network graph. This simple post-processing step can remove scattered errors and make the output labels more consistent.  

In the MRF post-processing approach, we use a pairwise term in the energy function of MRF to encourage the road segments, which are connected and \textit{belonging to the same logical road}, to have the same label. The energy function of the post-processing MRF is,
\begin{equation}
    E(x) = \sum_i -\log P(x_i) + \lambda \sum_{\text{connected } i,j} |x_i - x_j|^n
    \label{mrf_energy}
\end{equation}
We find the best hyper-parameters of MRF ($n$ and $\lambda$) through brute-force search on the validation set. At the inference time, we use belief propagation to minimize the energy function $E(x)$.

We also evaluate the CNN image classifier with larger receptive fields. The original CNN receptive field at each vertex is a 48x48 meter tile. We derive new baseline approaches by enlarging the receptive field to a 72x72 meter tile (1.5x) and a 96x96 meter tile (2.0x).

\subsection{Evaluation on Real-World Dataset}

\begin{table*}[ht]
	\centering
	\begin{tabular}{ |l||c|c||c|c||c|c| } 
		\hline
		Schemes &\# of Lane Acc.&Gain&Road Type Acc.&Gain& ALE &Reduction  \\ \hline 
		CNN Image Classifier \textbf{(naive baseline)} &71.8\%   & -  &89.1\%   & -  &0.374& -      \\ \hline 
		- with smoothing post-processing &\textbf{74.1\%} & \textbf{2.3\%}  & 90.6\%   & 1.5\% &\textbf{0.337}& \textbf{9.8\%} \\ \hline
		- with MRF post-processing &73.7\% &1.9\% &92.2\% & 3.1\% &0.355 & 5.1\% \\ \hline
		CNN Image Classifier (1.5x receptive field) &71.8\% & 0.0\%   &90.1\%   & 1.0\%   &0.367&1.9\%       \\ \hline 
		- with smoothing post-processing &74.0\% & 2.2\% &91.1\% &2.0\% &0.340 & 9.1\% \\ \hline
		- with MRF post-processing &\textbf{74.1\%} & \textbf{2.3\%}  &\textbf{92.9\%} & \textbf{3.8\%} &0.340 & 9.1\% \\ \hline
		CNN Image Classifier (2.0x receptive field) &68.8\%   & -2.0\%   &89.1\%   & 0.0\%   &0.393& -5.1\%       \\ \hline 
		- with smoothing post-processing &70.6\% & -1.2\% &89.9\% & 0.8\% &0.371 & 0.8\% \\ \hline
		- with MRF post-processing &70.2\% & -1.6\% &91.6\% & 2.5\% &0.386 & -3.2\% \\ \hline \hline
		\textbf{\name\ (ours)} &\textbf{77.2\%}   & \textbf{5.4\%}  &\textbf{93.1\%}   & \textbf{4.0\%}  &\textbf{0.291}& \textbf{22.2\%}       \\ \hline
	\end{tabular}
	\caption{Performance of \name\ and different CNN image classifier baselines. In the table, we highlight both the best and the second best results.}
	\label{table:different_graph_structures}
\end{table*}

We use the overall accuracy as metrics for both the number of lane prediction and the road type prediction. For the number of lanes prediction, a two-lane road may be incorrectly recognized as a three-lane road or even a six-lane road. However, the overall accuracy metric doesn't penalize more for the wrong prediction with six lanes than the wrong prediction with three lanes. Thus, we use an additional metric, the \textit{absolute lane error (ALE)}, to take the degree of error into account. We represent the output prediction of vertex $v$ as $y_v$ and the corresponding ground truth is $\hat{y_v}$,  where both $y_v$ and $\hat{y_v}$ are integers between $1$ and $6$. Then, the ALE is defined as, 
\begin{equation}
\text{ALE} = \frac{1}{|V|} \sum_{v \in V} |y_v - \hat{y_v}|
\end{equation}
We use this absolute lane error (ALE) as a complement to the overall accuracy metric in our evaluation.

\subsubsection{Comparison against baselines}

We report the overall accuracy of the two types of road attributes and the absolute lane error for the CNN image classifier baselines and \name\ in table \ref{table:different_graph_structures}. We show the result of \name\ with its best configuration in this table. As shown in the table, \name\ surpasses all the CNN image classifier based baselines. Compared with the baseline using only CNN image classifier, \name\ improves the inference accuracy of the number of lanes from 71.8\% to 77.2\%, and of the road type from 89.1\% to 93.1\%. This improvement comes with a reduction of the absolute lane detection error of 22.2\%. Compared with the best baseline (with MRF post-processing and 1.5x larger receptive field), \name\ still improves the accuracy of the lane count inference by 3.1 points, which comes with a reduction of the absolute lane detection error of 14.4\%, and achieves similar accuracy in road type inference. 

We show output examples of the number of lane prediction and the road type prediction in Figure \ref{fig:DeepRoadGraphExample}. We find \name\ performs more robust in many challenging places than the CNN image classifier. This is because the usage of the graph neural network enables \name\ to transitively incorporate information from  nearby road segments. Meanwhile, during training, unlike the CNN image classifier, \name\ treats all vertices on the sub-graph as a whole rather than treating each vertex independently. This doesn't force \name\ to learn how to map the image of a building into a two-lane road when the building occludes the road in the training dataset. Instead, \name\ can learn a more generic inductive rule to understand the spatial correlations and effect of different visual features (e.g., bridges, trees, intersections, etc) on road attributes. Although post-processing approach can be applied to fix some of the scattered errors, e.g., example (e) and (f) in Figure \ref{fig:DeepRoadGraphExample}, it cannot fix errors which requires more context information such as the errors in examples (a-d) in Figure \ref{fig:DeepRoadGraphExample}. This limitation is due to the \textit{information barrier} induced by the separation of local classification and global inference.

\subsubsection{Comparison within \name}
\begin{table}[ht]
	\centering
	\begin{tabular}{ |l|c|c|c|} 
		\hline
		Scheme& \# of Lane & Road Type & ALE    \\ \hline
		\name\ with& - & - & -  \\ \hline
		- Raw& 74.0\% & 91.2\% & 0.332  \\ \hline
		- Road& 75.5\% & 92.3\% & 0.327  \\ \hline
		- Road(D) & 75.6\% & 92.0\% & 0.324  \\ \hline
		- Raw+Road(D)+Aux& 77.2\% & 93.1\% & 0.291  \\ \hline
	\end{tabular}
	\caption{Impact of different graph structures used in \name. Here, we use abbreviations to denote different graphs. We use \textbf{Raw} for the original road network graph, \textbf{Road} for the road extraction graph, \textbf{Road(D)} for the road extraction graph with directional decomposition and \textbf{Aux} for the auxiliary graph for parallel roads.}
	\label{table:graph_structure}
\end{table}

Within \name, we first compare the performance of \name\ with different graph structures. We show results of \name\ with different graph structures in Table \ref{table:graph_structure}.

For a single graph structure, we find adding more restrictions into the graph structure can yield better performance, e.g., the performance of using road extraction graph is better than the performance of using the original raw road network graph. This is because in the road extraction graph, message propagation in the graph neural network is restricted to be within each logical road. This can remove the ambiguity of message propagation at intersections in the original road network graph, thus, improve performance. 

\name\ supports using multiple graph structures. We find using the combination of the Raw graph, Road(D) graph and Aux graph can yield better performance compared with the performance of using a single graph structure. This is because using multiple graph structures allows our neural network model to learn the best message propagation graph(s) for different attributes end-to-end.

As we mentioned before, we adopt two training techniques to improve the performance of \name. We show the comparison results in Table \ref{table:training_tech}. We find both of these two techniques are critical to the performance improvement of \name; the random vertex dropout has more impact on the number of lane inference and the graph Laplace regularization has more impact on the road type inference. 
\begin{table}
	\centering
	\begin{tabular}{ |l|c|c|c|} 
		\hline
		Scheme& \# of Lane & Road Type & ALE    \\ \hline
		\name&77.2\%& 93.1\% & 0.291  \\ \hline
		No Vertex Dropout& 74.7\% & 92.7\% & 0.325  \\ \hline
	    No Regularization.& 76.5\% & 90.8\% & 0.300  \\ \hline
	\end{tabular}
	\caption{Impact of random vertex dropout and graph Laplace regularization.}
	\label{table:training_tech}
\end{table}
\subsection{Evaluation on Synthesized Micro-Benchmark}

\begin{figure}
	\begin{center}
		\includegraphics[width=0.95\linewidth]{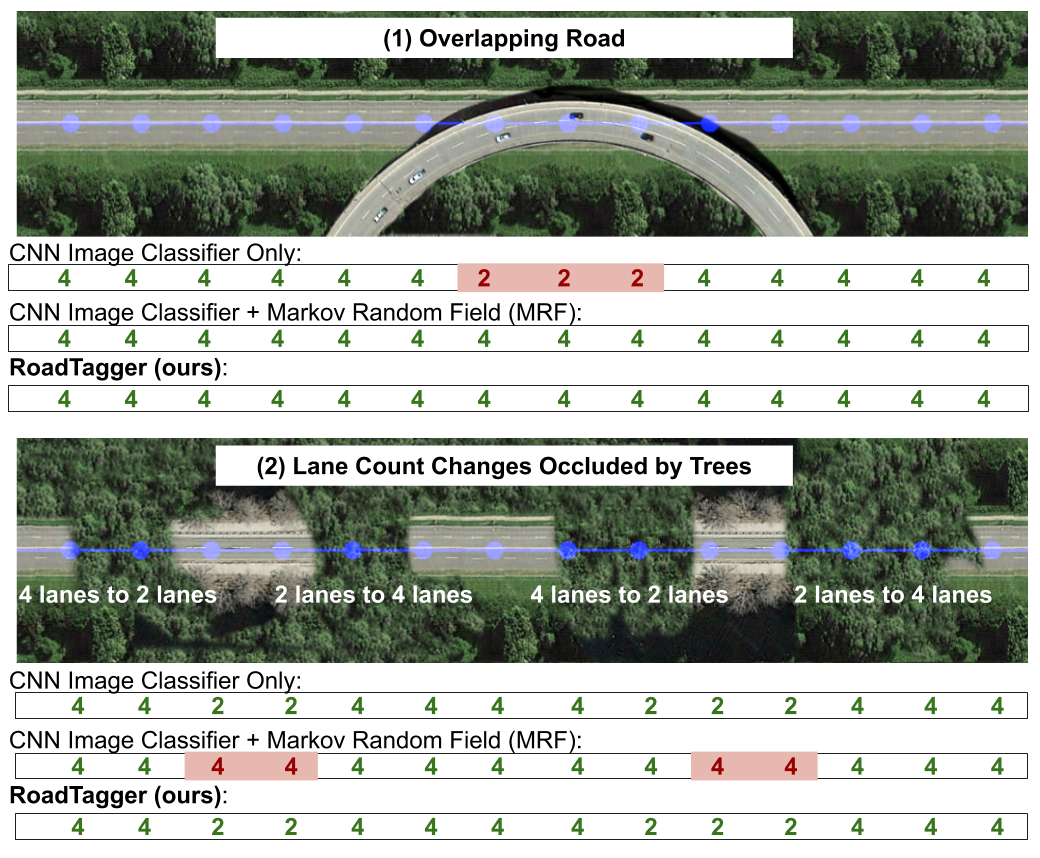}
	\end{center}
	\caption{Two representative samples of the micro benchmark. \name\ predicts both of them correctly.}
	\label{fig:microbenchmark}
\end{figure}

We conduct an extensive evaluation of \name\ on a micro benchmark. In this micro benchmark, we inject different types of challenges to the satellite imagery. We would like to study the impact of occlusions with different types and amounts as well as other challenges such as missing lane markings in a controlled way. In this micro benchmark, we find \name\ is robust to a wide range of different disruptions. These disruptions include removing all the lane markings on part of the roads, alternatively occluding the left and right side of the roads, and even occluding the target road with an overpass road.

We show two representative examples of this benchmark in Figure \ref{fig:microbenchmark}. \textit{Please refer to the supplementary material for the evaluation on the whole micro-benchmark.}

We find the two examples shown in Figure \ref{fig:microbenchmark} particularly interesting. In example (a), an overpass road occluded the target road. In example (b), the lane count changes when the road is occluded by trees. To correctly predict the number of lanes in both example (a) and (b), \name\ needs to know that when the starting point of an overpass road is detected, the visual features of the overpass should be ignored until the far edge of the overpass is detected. At the same time, \name\ needs to know if the road is temporarily occluded by trees, the following road segment still belongs to the same target road. We find \name's end-to-end architecture enables it to learn all this knowledge correctly without any additional labels or explicit features. This is perhaps the most attractive part of \name. 

\section{Discussion}
\textbf{Can \name\ Generalize to City-Scale Graphs?} In our evaluation, we find \name\ can generalize well to city-scale graphs. During inference, \name\ labels the whole road network graph (with 3,000 to 4,000 vertices) for each 2km by 2km region in one shot (Use the entire road network graph as input). We find training \name\ with 256-node subgraphs can generalize well in larger graphs, e.g., graphs with 3,000 to 4,000 nodes.

\textbf{Errors Made by \name.} We observe two types of errors made by \name\ in our evaluation. (1) We find \name\ makes wrong predictions for invisible roads (occluded by trees or buildings) when the disruptions are longer than the GNN propagation step (we show examples of this type of failure in the supplementary material). We think this type of error can be eliminated through enlarging the propagation step and the subgraph size (i.e., 256) during training. (2) We find \name\ outputs road attributes in ABABA style along the road when the road attribute is ambiguous. We think this issue can be addressed by incorporating GAN into \name\ framework.

\section{Conclusion}
In this paper, we propose \name. \name\ adapts a novel combination of CNN and graph neural network to enable end-to-end training for road attribute inference. This framework eliminates fundamental limitations of the current state-of-the-art that relies on a single CNN with post-processing. 
We conduct a comprehensive evaluation of the performance and robustness of \name; the evaluation result shows a significant improvement in both performance and robustness compared with the current state-of-the-art. The result also shows \name's strong inductive reasoning ability learned end-to-end. We believe \name\ framework is a fundamental improvement in road attributes inference and can be easily extended to other road attributes.


\bibliographystyle{aaai}
\bibliography{egbib}

\end{document}